\pdfoutput=1

\documentclass[11pt,table]{article}

\usepackage{EACL2023}
\usepackage{booktabs}
\usepackage{amsmath}
\usepackage{times}
\usepackage{latexsym}
\usepackage{float}
\usepackage[export]{adjustbox}
\usepackage{topicvis}

\usepackage[T1]{fontenc}

\usepackage[utf8]{inputenc}

\usepackage{microtype}

\usepackage{inconsolata}

\title{Dynamic embedded topic models and change-point detection for exploring literary-historical hypotheses}

\author{Hale Sirin \\
  Center for Digital Humanities\\
  Johns Hopkins University\\
  \texttt{hsirin1@jhu.edu} \\\And
  Tom Lippincott \\
  Center for Digital Humanities\\
  Johns Hopkins University\\
  \texttt{tom.lippincott@jhu.edu} \\}

\begin{document}
\maketitle
\begin{abstract}
We present a novel combination of dynamic embedded topic models and change-point detection to explore diachronic change of lexical semantic modality in classical and early Christian Latin. We demonstrate several methods for finding and characterizing patterns in the output, and relating them to traditional scholarship in Comparative Literature and Classics. This simple approach to unsupervised models of semantic change can be applied to any suitable corpus, and we conclude with future directions and refinements aiming to allow noisier, less-curated materials to meet that threshold.
\end{abstract}

\section{Introduction}
Characterizing and interpreting linguistic novelty has a long tradition in both humanistic scholarship. A foundational study in comparative literature \cite{figura} hinges empirically on shifts in the meanings of particular words, most notably \emph{figura}. He claims that \emph{figura} went from particularly abstract (as a translation of Plato's \emph{schema}), to concrete (due in part to several particularly novel authors), and finally was usable in both senses in the writing of early Church fathers. We refer to this type of semantic shift as \emph{bimodality}, the degree to which a word makes a sharp transition between having one or two senses.\footnote{For simplicity we limit this study to \emph{bi}modality, and leave higher complexity to future work.} In our attempts to reproduce and extend this humanistic hypothesis, we make the following contributions:

\begin{itemize}
\item Propose a novel combination of unsupervised machine learning methods for surfacing relevant phenomena
\item Demonstrate viewpoints on model output that move readily between general trends and specific observations
\item Derive legible humanistic insights and lines of inquiry regarding shifts in Latin through the Classical and early Christian periods
\end{itemize}

\section{Background}

Our goals and methods in this paper have connections to research tracking semantics across time in embedding spaces \cite{emb}, which we differ from in our focus on modality rather than the geometric position. Also related is the long-standing task of word sense disambiguation (WSD) \cite{wsd}, which we differ from by not operating with a gold standard sense inventory to target, or other supervised task.

The dynamic embedded topic model (DETM) \cite{detm} extends topic models \cite{tm} to operate over word embeddings, and capture topic evolution over time. Change-point detection \cite{pelt} considers the problem of determining if and where the distribution generating a sequence of observations changes, typically w.r.t. time. The simplest approach, employed here, uses dynamic programming to find the optimal piecewise-linear fit to the observations. The Jensen-Shannon divergence (JSD) \cite{jsd} is a symmetric distance measure based on the Kullback-Leibler divergence.

The Perseus project \cite{perseus} is a long-standing database and interface to a curated corpus of primary sources from the Classical and early Medieval world.

\section{Materials and methods\footnote{Code from the study is available at \url{https://github.com/comp-int-hum/diachronic-latin}}}

We derive our corpus from the Perseus project by extracting all XML documents from the underlying repositories, and extract all text marked as \emph{Latin} along with the name of the purported author.  We then manually assign years to the set of unique authors based on rough scholarly consensus, and keep materials that fall between 250 BCE and 500 CE. This leads to a corpus of 574 documents from 101 authors. We use the Classical Language Toolkit \cite{cltk} to lemmatize each token and filter non-Latin vocabulary. We group documents into 75-year windows, and split documents into sub-documents of at most 500 tokens.

\begin{table}[H]
\tiny
\centering
\begin{tabular}{llll}
\toprule
Word & \multicolumn{3}{c}{Neighbors} \\
\midrule
bellum & proelium:0.53 & optatus:0.49 & bello:0.46 \\
hasta & clipeus:0.46 & sarisa:0.46 & tragula:0.44 \\
terra & caelum:0.52 & introgredior:0.50 & inhabitabilis:0.50 \\
ignis & flamma:0.55 & exuro:0.52 & ardeo:0.51 \\
debeo & cumulatus:0.57 & oppignero:0.56 & faeneratio:0.54 \\
\bottomrule
\end{tabular}
\caption{Nearest neighbors of several words in the initial word2vec embedding space.}
\label{table:nns}
\end{table}

We initialize a 50-topic DETM with skip-gram embeddings \cite{word2vec} trained on the corpus.  Table \ref{table:nns} shows the nearest neighbors for several words, to demonstrate the intuitiveness of the initial embedding space. We fit the DETM using the hyper-parameters listed in Appendix \ref{sec:hypers}, monitoring perplexity on the dev set for learning rate adjustment and early termination.

\subsection{Measuring static and diachronic semantics}

We define a word's \emph{bimodality}, within a particular window, as the degree to which its probability mass is evenly and exhaustively between two topics.  At each time window and for each word, we use the two highest values from the word's empirical distribution over topics, \( first \) and \( second \), to compute a score:

\begin{align*}
evenly\_distr &= 1.0 - (first - second) \\
exhaustive &= first + second \\
bimodality &= \frac{evenly\_distr + exhaustive}{2}
\end{align*}

This \emph{bimodality} takes its maximal value of \( 1.0 \) when the word is evenly split between two topics. Using the word's sequence of bimodality scores, we apply change-point detection with an L2 cost to find the window constituting the most-prominent shift in modality, which we refer to as the word's \emph{change-point}.  Finally, we compute the absolute value of the difference between the means on each side of the change-point, which we refer to as the word's \emph{delta}.

We define the \emph{novelty} of an author as the degree to which their topic distribution diverges from that of the time window immediately preceding their own. For this we calculate the Jensen-Shannon divergence. Note that, while deltas and novelties are derived from the same model output and aren't independent, they can provide different useful perspectives on change, as our results show.

\section{Results and analysis}

\begin{figure}[H]
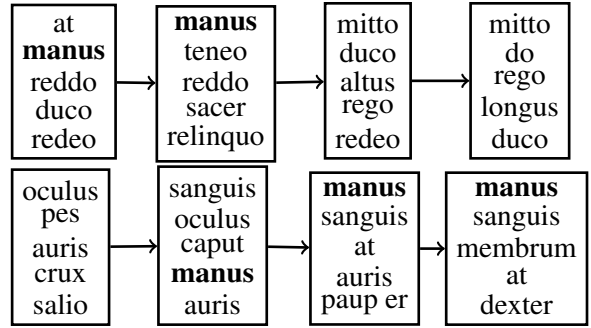

\topicevolution{21}{0}{0}{{{at,\textbf{manus},reddo,duco,redeo},{\textbf{manus},teneo,reddo,sacer,relinquo},{mitto,duco,altus,rego,redeo},{mitto,do,rego,longus,duco}}}
\topicevolution{20}{0}{0}{{{oculus,pes,auris,crux,salio},{sanguis,oculus,caput,\textbf{manus},auris},{\textbf{manus},sanguis,at,auris,paup
er},{\textbf{manus},sanguis,membrum,at,dexter}}}
\caption{Top words of two topics, at four windows evenly spread across our temporal range, illustrating the semantic shift of \emph{manus} (\emph{hand}).}
\label{fig:evo}
\end{figure}

To exemplify the phenomenon of interest, and as an initial qualitative example, Figure \ref{fig:evo} places snapshots of two topics side-by-side. The first topic focuses on actions (commanding, holding, sending, ruling), the second on body parts (eye, head, limb, ear). The word \emph{manus} (\emph{hand}) moves from the former to the latter, which we interpret as a shift from a figurative to corporeal sense. The overlap in the second window corresponds to bimodality. Emergent examples like this lend credibility to our approach as we aggregate and look for broader patterns.

\begin{figure}[H]
\includegraphics[width=.45\textwidth]{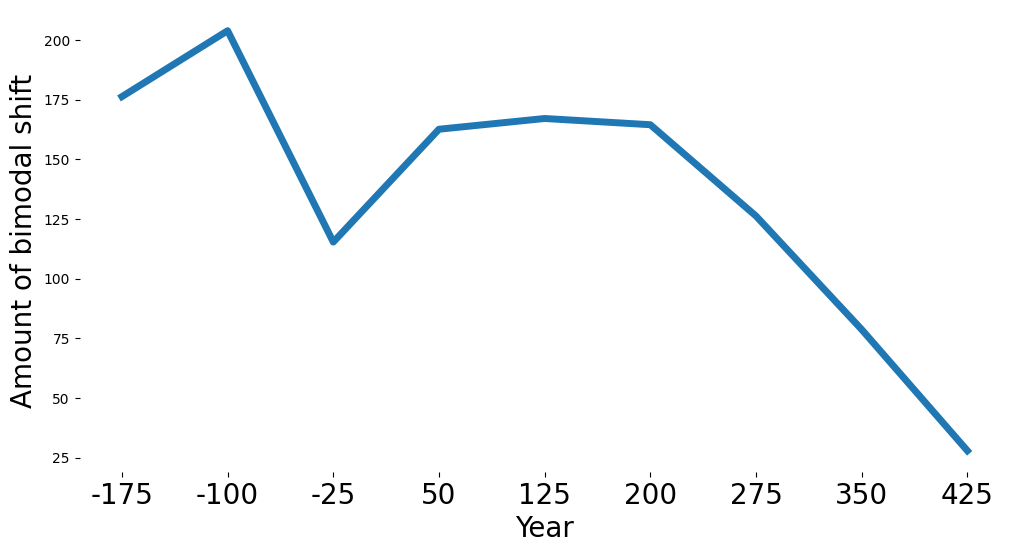}
\caption{Sum of deltas (bimodal shift) for words with their change-point in the given window.}
\label{fig:change-point}
\end{figure}

Our highest-level view is from summing, in each window, the deltas from all change-points identified therein.  Figure \ref{fig:change-point} plots these sums across time: the overarching trend is a substantial decrease in modality shift starting around 200CE with the early Christian era.

\begin{figure}[h]
\includegraphics[width=.45\textwidth]{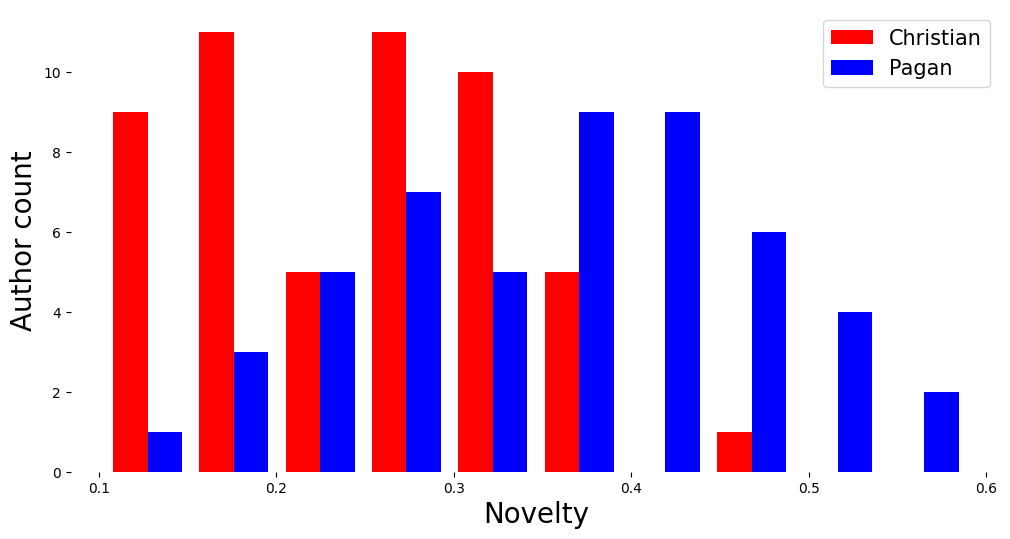}
\caption{Counts of Christian and pagan authors binned into 10 ranges according to their novelty.}
\label{fig:histogram}
\end{figure}

Figure \ref{fig:histogram} demonstrates that this is not merely a broader linguistic trend projected onto the increasing dominance of Christian writers.  The lowest-novelty non-Christian (leftmost blue) is Terentius Afer, one of the earliest writers in the data, while the most-novel Christian (rightmost red) is Saint Hilary, who falls in the middle of the Christian range.  This is the opposite of the expected outcome if Latin had simply undergone a general reduction in novelty over time.  It supports the view that the Christian writers were, intentionally or naturally, standardizing their language along with their religion.

\begin{table}[H]
\small
\centering
\begin{tabular}{lrl}
\hline
Author & Window & JSD \\
\hline
Apicius & 425 & 0.589 \\
Vitruvius & -100 & 0.553 \\
Vergil & -100 & 0.526 \\
Julius Caesar & -100 & 0.507 \\
Musonius Rufus & 50 & 0.506 \\
\ldots & \\
Terence & -175 & 0.126 \\
\emph{Rufinus of Aquileia} & 350 & 0.124 \\
\emph{Hegemonius} & 350 & 0.122 \\
\emph{Augustine} & 350 & 0.106 \\
\emph{Saint Jerome} & 350 & 0.103 \\
\hline
\end{tabular}
\caption{The most and least novel authors in the corpus and their temporal window, according to JSD between their topic distributions and the topic distribution of the preceding temporal window.  Early Christian writers are italicized.}
\label{table:authors}
\end{table}

\begin{figure}[h]
\includegraphics[width=.45\textwidth]{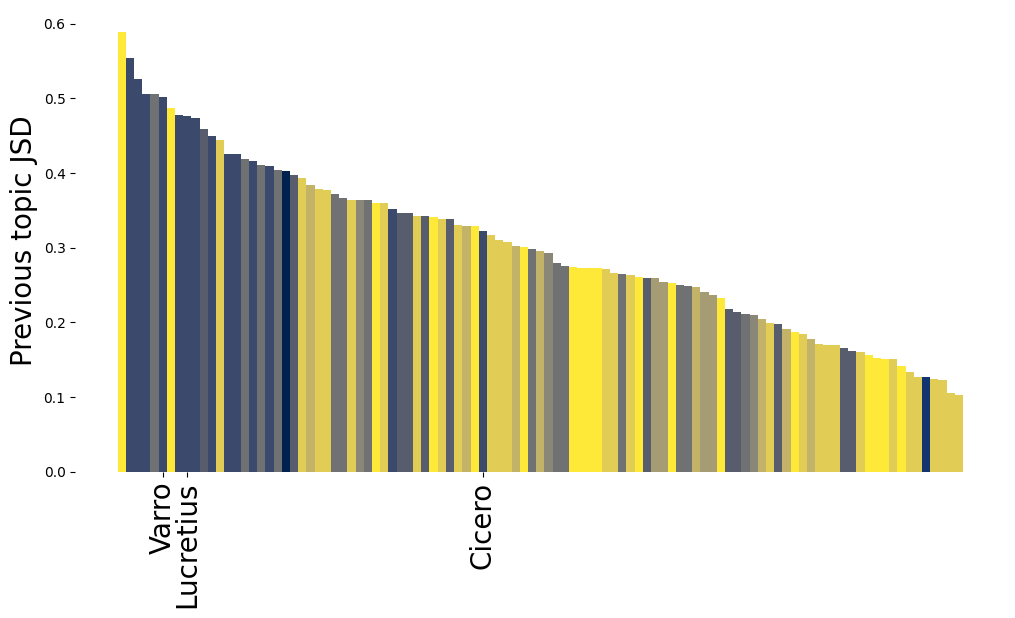}
\caption{All author novelties in descending order, indicating the position of several authors singled out by Auerbach. Darker colors correspond to earlier windows.}
\label{fig:authors}
\end{figure}

The five most and least-novel authors are shown in Table \ref{table:authors}. The most striking pattern is the dominance of early Christian authors as the least-novel: this supports the trend seen in Figure \ref{fig:change-point}. The one non-Christian author in the bottom five, Terence, is an interesting case: as a former slave from North Africa one might expect his writing to be rather novel, but his position in our results might align with the common view that his Latin is particularly clear and standard, or be affected by data sparsity in the preceding time period (the earliest in our corpus).

The most-novel authors often focus on a unique domain: Apicius is the (likely composite, \citet{vehling:1936}) author of a recipe collection, while Vitruvius produced the first technical treatise on architecture. It's unsurprising that specialized domains lead to outliers, while Vergil and Caesar may be unsurprising for narrative and stylistic properties.  To our knowledge Rufus (a philosopher of the early empire) has not before been highlighted as particularly distinctive, and so might be a compelling target for closer analysis.

Earlier authors have higher novelty, as shown by the darker lefthand columns in Figure \ref{fig:authors}, and aligning with the trend in word deltas.  Of authors Auerbach considered novel w.r.t. \emph{figura}, Lucretius and Varro indeed fall in the top range of novelty, while Cicero is only slightly above the median (in fact, he falls lower than his less-famous brother).

\begin{table}[H]
\small
\centering
\begin{tabular}{lrl}
\hline
Word & Year & Delta \\
\hline
cathedra (\emph{chair}) & -175 & 0.947 \\
cicatrix (\emph{scar}) & 425 & 0.944 \\
conlatio (\emph{bring together}) & 350 & 0.939 \\
auster (\emph{south wind}) & 350 & 0.927 \\
recte (\emph{upright}) & 350 & 0.915 \\
\ldots & & \\
probo (\emph{make good}) & 350 & 0.004 \\
obsidio (\emph{siege}) & -25 & 0.004 \\
sollicitudo (\emph{anxiety}) & 350 & 0.003 \\
declaro (\emph{disclose}) & 350 & 0.002 \\
corpus (\emph{body}) & 125 & 0.001 \\
\hline
\end{tabular}
\caption{Words whose primary change-point divides their modality with the greatest and least deltas.  The year is the start of the change-point's window, with negative numbers corresponding to BCE.}
\label{table:words}
\end{table}

\begin{figure}[h]
\includegraphics[width=.45\textwidth]{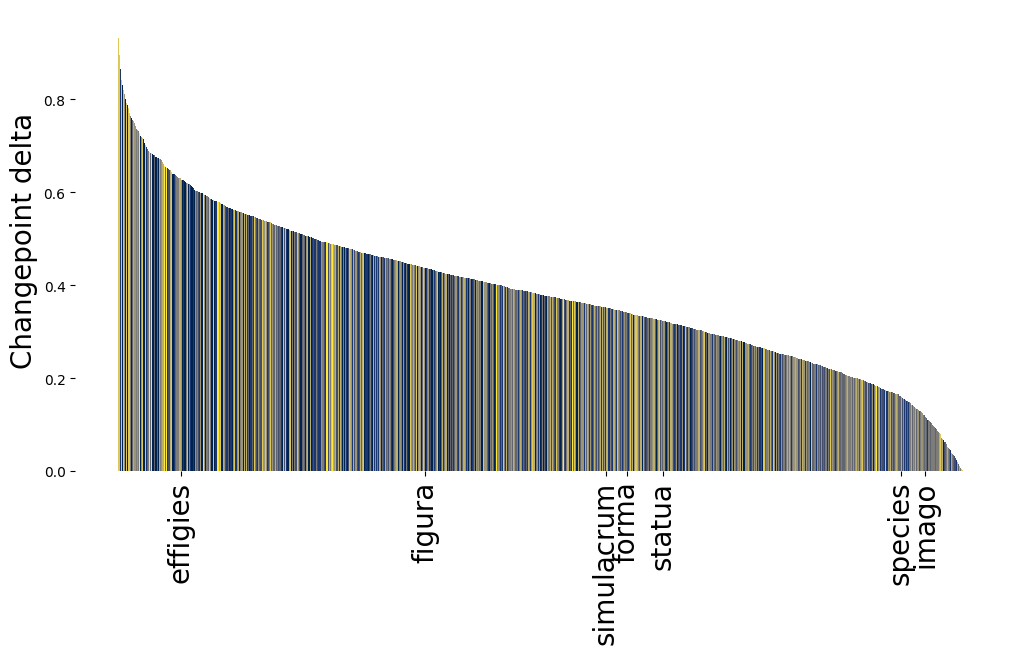}
\caption{Words sorted by degree of bimodal shift (their change-point delta).}
\label{fig:words}
\end{figure}

The five words with the highest and lowest deltas, shown in Table \ref{table:words}, contrasts with the pattern of decreasing change in summed deltas and author novelty. Most of the top words are excellent examples of the early Christian church's adaptation of Classical vocabulary. \emph{Cicatrix} (\emph{scar}), for instance, takes on figurative meaning when speaking of Christ's wounds as a portal to salvation. \emph{Auster} (\emph{south wind}) may have taken a similar turn: Cicero makes the poetic, but grounded, statement that his ship was carried back to Rome by the south wind, while Augustine compares the Holy Spirit's wrath against wrongdoers to the south wind scattering dust. \emph{Conlatio} (alternate form of \emph{collatio}, \emph{bring together, unify}), and \emph{recte} (\emph{upright, vertical, well-guided}) seem intuitive shifts for the early Christian era. \emph{Cathedra} (\emph{chair, office}), whose 175BCE change-point comes before it is well-attested, may be an artifact of high variance: an important future goal is to leverage data sparsity information within the modeling process, ideally to produce a measure of robustness for change-points.

When the vocabulary is arranged by decreasing delta, we can inspect the positions of several words from Auerbach's study of \emph{figura}. Interestingly, \emph{figura} is the second-highest of the terms, a considerable distance below the high-delta \emph{effigies}.

\begin{figure}[h]
\includegraphics[width=.45\textwidth]{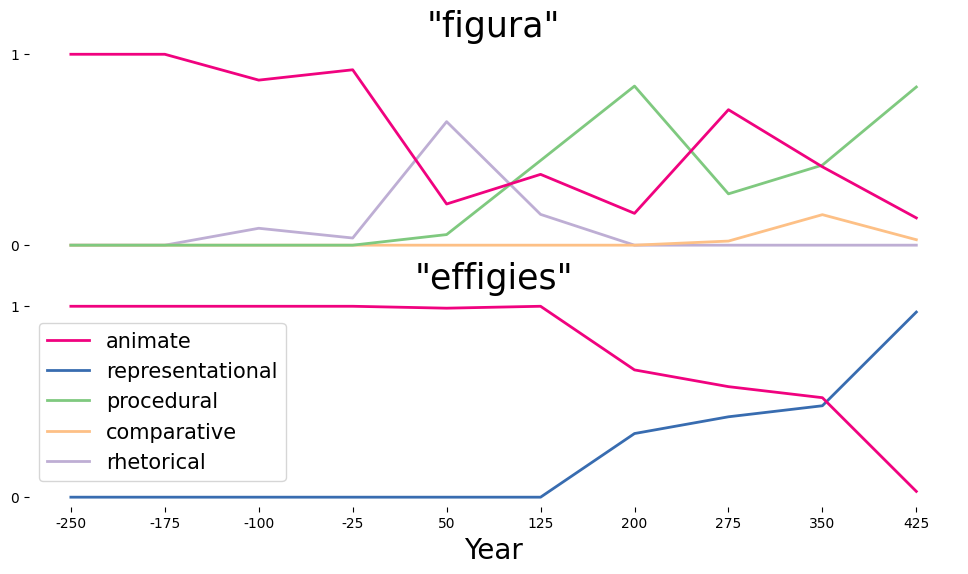}
\caption{The temporal evolution of topics responsible for the words \emph{figura} (\emph{form}, \emph{shape}) and \emph{effigies} (\emph{copy}, \emph{imitation}), with provisional characterizations of meaning. Both words are initially dominated by the \emph{tangible} topic.}
\label{fig:detail}
\end{figure}

Inspecting the two words more closely, Figure \ref{fig:detail} shows their empirical topic distributions over time. Both words are initially unimodal, generated from the same topic that focuses on animate nouns, particularly the human body, but they evolve quite differently:

\emph{effigies} is the simpler case: it remains unimodal in the animate topic until about 125CE, at which point it begins shifting to a topic focused on representing, referring, imitating, and equating. After 300 years, \emph{effigies} is again unimodal, but with respect to this \emph{representational} topic.

\emph{figura} is more complex: rather than directly swapping the animacy topic for another monotonically, it fluctuates between three other topics that themselves are somewhat dynamic. The \emph{rhetorical} spike involves vocabulary related to argumentation and speaking.\footnote{An interesting topic in its own right, it seems to temporally proceed from a focus on learning and understanding, to a focus on the tension between groups and individuals.} The later \emph{comparative} spike has vocabulary used to establish (particularly, temporal and spatial) relationships. The largest shift, however, is to the \emph{procedural} topic, which ends our time-frame as the dominant sense. It is also the most difficult to interpret: the gloss was chosen because of the unusual strength of verbs denoting a change of state or coming-to-be (proceed from, generate, come forth, burst, grow), and related "source" nouns (seed, fountain, sea, earth). This final sense may come to dominate due to the expanding use of Christian idioms ("brought forth on the earth", etc), but is well-attested throughout the Classical era.

Taken together, these different viewpoints may begin to disentangle two distributional shifts: one an acute, limited adaptation in a small number of lexical items in the early Christian era, the other a longer, more diffuse process that appears to be slowing down over the same time period.

Unfortunately, the vast majority of scholarship regarding liturgical language is born in, and concerned with, theology. Exceptions, e.g. \citet{liddicoat:1993}, do highlight the critical early need to \emph{define orthodoxy} and then \emph{create stability}.  These two concerns map to the two distributional shifts highlighted above: focused modification of specialized vocabulary, and broader linguistic consistency to consolidate the early Church.

\section{Future Work}
Deeper scrutiny of model output, involving scholars from Classics, would benefit from \emph{in situ} examples drawn automatically from the underlying sources. Having established its ability to surface historically distinctive authors and vocabulary, we are augmenting the pipeline in this direction, in anticipation of implementing a frontend for humanists to apply and explore their own diachronic corpora.

A critical facet of Auerbach's arguments is the permeability and comingling of the languages, and a suitable Greek lexicon and lemmatizer would make it a straightforward to include prior and contemporary Greek writing.

The greatest barrier to extending our methodology to arbitrary languages and time periods is imperfect and low-coverage data. In parallel with this research we have applied the same method to a noisy Latin corpus derived from the HathiTrust \cite{htc}, which offers considerably higher coverage of sources. Unfortunately the level of noise (from OCR, commentary, etc) and redundancy renders it challenging to use credibly without extensive post-processing. We plan to use this research as a case study for developing a core set of methods for gathering, deduplicating, and rectifying an arbitrary HTC-based corpus such that it approaches the fidelity of a manually-curated resource like Perseus.
\newpage

\bibliography{diachronic_latin.bib}

\begin{thebibliography}{13}
\expandafter\ifx\csname natexlab\endcsname\relax\def\natexlab#1{#1}\fi

\bibitem[{Auerbach(1959)}]{figura}
Erich Auerbach. 1959.
\newblock \emph{{Scenes from the Drama of European Literature: six essays}}.
\newblock Meridian Books.

\bibitem[{Blei et~al.(2003)Blei, Ng, and Jordan}]{tm}
David~M. Blei, Andrew~Y. Ng, and Michael~I. Jordan. 2003.
\newblock {Latent Dirichlet Allocation}.
\newblock \emph{J. Mach. Learn. Res.}, 3(null):993–1022.

\bibitem[{Crane(2023)}]{perseus}
Gregory~R. Crane. 2023.
\newblock \href {http://www.perseus.tufts.edu} {{Perseus Digital Library}}.

\bibitem[{Dieng et~al.(2019)Dieng, Ruiz, and Blei}]{detm}
Adji~B. Dieng, Francisco J.~R. Ruiz, and David~M. Blei. 2019.
\newblock \href {http://arxiv.org/abs/1907.05545} {{The Dynamic Embedded Topic
  Model}}.

\bibitem[{Hamilton et~al.(2016)Hamilton, Leskovec, and Jurafsky}]{emb}
William~L. Hamilton, Jure Leskovec, and Dan Jurafsky. 2016.
\newblock \href {http://arxiv.org/abs/1605.09096} {{Diachronic Word Embeddings
  Reveal Statistical Laws of Semantic Change}}.
\newblock \emph{{CoRR}}, abs/1605.09096.

\bibitem[{HathiTrust Foundation, 2023()}]{htc}
HathiTrust Foundation, 2023. 2023.
\newblock \href {https://www.hathitrust.org} {{HathiTrust Digital Library}}.

\bibitem[{Ide and V{\'e}ronis(1998)}]{wsd}
Nancy Ide and Jean V{\'e}ronis. 1998.
\newblock {Introduction to the special issue on word sense disambiguation: the
  state of the art}.
\newblock \emph{{Computational linguistics}}, 24(1):1--40.

\bibitem[{Johnson et~al.(2014--2021)Johnson, Burns, Stewart, and Cook}]{cltk}
Kyle~P. Johnson, Patrick Burns, John Stewart, and Todd Cook. 2014--2021.
\newblock \href {https://github.com/cltk/cltk} {{CLTK: The Classical Language
  Toolkit}}.

\bibitem[{Killick et~al.(2012)Killick, Fearnhead, and Eckley}]{pelt}
R.~Killick, P.~Fearnhead, and I.~A. Eckley. 2012.
\newblock \href {https://doi.org/10.1080/01621459.2012.737745} {{Optimal
  Detection of Changepoints With a Linear Computational Cost}}.
\newblock \emph{{Journal of the American Statistical Association}},
  107(500):1590–1598.

\bibitem[{Liddicoat(1993)}]{liddicoat:1993}
Anthony~J. Liddicoat. 1993.
\newblock \href {https://doi.org/https://doi.org/10.1075/aral.16.2.06lid}
  {Choosing a liturgical language: Language policy and the catholic mass}.
\newblock \emph{Australian Review of Applied Linguistics}, 16(2):123--141.

\bibitem[{Lin(1991)}]{jsd}
J.~Lin. 1991.
\newblock \href {https://doi.org/10.1109/18.61115} {{Divergence measures based
  on the Shannon entropy}}.
\newblock \emph{{IEEE Transactions on Information Theory}}, 37(1):145--151.

\bibitem[{Mikolov et~al.(2013)Mikolov, Chen, Corrado, and Dean}]{word2vec}
Tomas Mikolov, Kai Chen, Greg Corrado, and Jeffrey Dean. 2013.
\newblock \href {http://arxiv.org/abs/1301.3781} {{Efficient Estimation of Word
  Representations in Vector Space}}.

\bibitem[{Vehling(1936)}]{vehling:1936}
Joseph~Dommers Vehling. 1936.
\newblock \emph{{Cooking and Dining in Imperial Rome}}.
\newblock Walter M. Hill.

\end{thebibliography}
\bibliographystyle{acl_natbib}

\appendix

\section{Hyper-parameters}
\label{sec:hypers}
\noindent
\begin{tabular}{lr}
\multicolumn{2}{c}{word2vec} \\
\hline
Name & Value \\
\hline
Method & skip-gram \\
Window size & 5 \\
Embedding size & 300 \\
Epochs & 10 \\
\hline
\end{tabular}

\vspace{1cm}
\noindent
\begin{tabular}{lr}
\multicolumn{2}{c}{Dynamic embedded topic model} \\
\hline
Name & Value \\
\hline
Topics & 50 \\
Epochs & 1000 \\
Batch size & 2000 \\
Learning rate & 0.016 \\
\hline
\end{tabular}

\section{Caveats}

We note that our aim in \emph{Results and analysis} is to illustrate productive exploratory methods and seed discussion with Classicists and literary theorists regarding if and how the patterns relate to traditional scholarship such as Auerbach's. Observations, and certainly interpretations, are provisional. Document dates were assigned as precisely as possible after a light survey, but most often are approximated as the midpoint of the author's life, or of the century they are believed to have flourished (the assignments are included in the experimental repository, for scrutiny and revision). Where English is used to characterize a topic, it is a provisional gloss of a complex, dynamic concept; where used to translate a Latin word, it is derived from the Lewis dictionary.
\end{document}